# Preoperative Volume Determination for Pituitary Adenoma

Dženan Zukić[a *], Jan Egger[b, c], Miriam H. A. Bauer[b, c], Daniela Kuhnt[b], Barbara Carl[b]
Bernd Freisleben[c], Andreas Kolb[a] and Christopher Nimsky[b]

[a] University of Siegen, Computer Graphics Group, Hölderlinstrasse 3, 57076 Siegen, Germany;
[b] University of Marburg, Department of Neurosurgery, Baldingerstrasse, 35033 Marburg, Germany;
[c] University of Marburg, Department of Mathematics and Computer Science, Hans-Meerwein-Str. 3, 35032 Marburg, Germany;

**ABSTRACT**

The most common sellar lesion is the pituitary adenoma, and sellar tumors are approximately 10-15% of all intracranial neoplasms. Manual slice-by-slice segmentation takes quite some time that can be reduced by using the appropriate algorithms. In this contribution, we present a segmentation method for pituitary adenoma. The method is based on an algorithm that we have applied recently to segmenting glioblastoma multiforme. A modification of this scheme is used for adenoma segmentation that is much harder to perform, due to lack of contrast-enhanced boundaries. In our experimental evaluation, neurosurgeons performed manual slice-by-slice segmentation of ten magnetic resonance imaging (MRI) cases. The segmentations were compared to the segmentation results of the proposed method using the Dice Similarity Coefficient (DSC). The average DSC for all datasets was 75.92%±7.24%. A manual segmentation took about four minutes and our algorithm required about one second.

## 1. INTRODUCTION

Approximately 10-15% of all intracranial neoplasms are sellar tumors. The most common sellar lesion is the pituitary adenoma[1,2]. The lesions can be classified according to size or hormone-secretion (hormone-active and hormone-inactive). Microadenomas are less than 1 cm in diameter, whereas macroadenomas measure more than 1 cm. The rare giant-adenomas have more than 4 cm in diameter.

Secreted hormones can be cortisol (Cushing's disease), human growth hormone (hGH; acromegaly), follicle-stimulating hormone (FSH), luteinising hormone (LH), thyroid-stimulating hormone (TSH), prolactine, or a combination of these. Only for the prolactine-expressing tumors, a pharmacological treatment is the initial treatment of choice in form of dopamine-agonists. Treatment is most commonly followed by a decrease of prolactine-levels and tumor volume. For acromegaly and Cushing's disease, surgery remains the first-line treatment, although somatostatin receptor analogues or combined dopamine/somatostatin receptor analogues are a useful second-line therapeutical option for hGH-expressing tumors. Current medical therapies for Cushing's disease primarily focus on the adrenal blockade of cortisol production, although pasireotide and cabergoline show promise as pituitary-directed medical therapy for Cushing's disease.

Thus, not only for the most hormone-active, but also for hormone-inactive macroadenomas with mass-effect, surgery is the treatment of choice, most possibly via a transsphenoidal approach[3]. For hormone-inactive mircroadenomas (<1cm) there is no need for immediate surgical resection. The follow-up contains endocrine and ophthalmological evaluation as well as magnetic resonance imaging (MRI). In case of continuous tumor volume progress, microsurgical excision is the treatment of choice. Thus, the tumor volume should be tracked over the time of the follow-up.

In this contribution, we present a segmentation method for pituitary adenomas. The method is based on an algorithm we recently developed for segmenting glioblastoma multiforme (GBM)[4].

The paper is organized as follows. Section 2 discusses related work. Section 3 presents the details of the proposed approach. Section 4 discusses experimental results and concludes the paper.

---

[*] Further author information: (Send correspondence to Dž. Z.)
Dž. Z.: E-mail: zukic@fb12.uni-siegen.de, Telephone: +49 271 740 2826
J. E.: E-mail: egger@med.uni-marburg.de, Telephone: +49 6421 58 66754

## 2. RELATED WORK

Several algorithms have already been proposed for segmenting brain tumors in MRI (magnetic resonance images). A relatively recent overview of some deterministic and statistical approaches is given by Angelini et al.[5]. Most of these approaches are region-based; more recent ones are based on deformable models and include edge-information.

Neubauer et al.[6] and Wolfsberger et al.[7] introduce STEPS, a virtual endoscopy system designed to aid surgery of pituitary tumors. STEPS uses a semi-automatic segmentation method that is based on the so-called watershed-from-markers technique. The watershed-from-markers technique uses manually defined markers in the object of interest and the background. A memory efficient and fast implementation of the watershed-from-markers algorithm – also extended to 3D – has been developed by Felkel et al.[8].

Descoteaux et al.[9] have proposed a novel multi-scale sheet enhancement measure and have applied it to paranasal sinus bone segmentation. The measure has the essential properties to be incorporated in the computation of anatomical models for the simulation of pituitary surgery.

Egger et al.[10] have recently presented a graph-based method for pituitary adenoma segmentation. The method starts by setting up a directed and weighted 3D graph from a user-defined seed point that is located inside the pituitary adenoma. To set up the graph, the method samples along rays that are sent through the surface points of a polyhedron with the seed point as the center. After graph construction, the minimal cost closed set on the graph is computed via a polynomial time s-t cut[11].

## 3. METHOD

The proposed method is initialized by an approximate outline on a slice near the center of the tumor, which is drawn by the user (Figures 1 and 2). From this user input the following initial information is obtained:

1. Center of the tumor: The X and Y coordinates are calculated as the center of boundary-enclosed area, and the Z coordinate is the index of the selected slice.

2. The minimum and maximum intensities of voxels of interest: Intensities of interest are all those within the boundary on the selected slice, ignoring the few highest and lowest percent in order to account for noise.

3. The average "radius": The average distance from the center to the boundary of a sphere-like object is a dimensionality-invariant measure – it is the same in 2D (slice) and 3D (whole volume), e.g. the radius of a circle is equal to the radius of the sphere made by rotating that circle around its diameter.

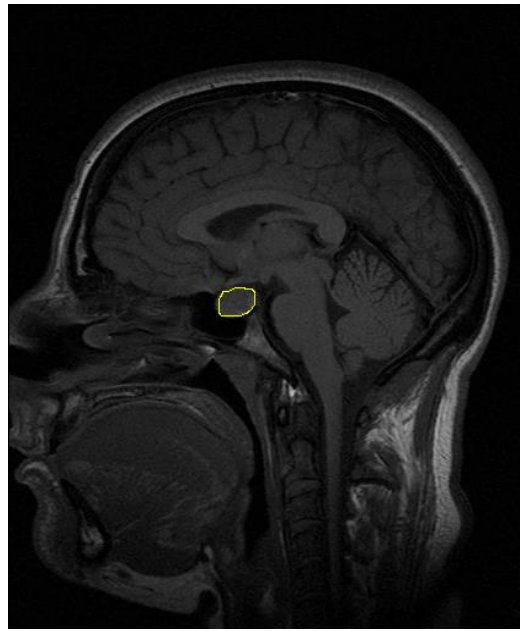

Fig. 1: An example of an initialization, shown in yellow.

The algorithm starts with a small triangular surface mesh, i.e. a triangulated cube that turns into a sphere after a few iterations, at the approximate center of the tumor. This mesh is enlarged using balloon inflation forces[12], enforcing a star-shaped† geometry. Since the vertices are moved only into regions with an intensity of interest, the mesh is not inflated beyond the glioma boundary. The segmentation is finished when the user-initialized "radius" is reached, or when the maximum number of iterations is exceeded (if the segmentation has converged sooner than estimated).

---

† A star-shaped object is an object in which a point (the center) exists that can be connected with every surface point by a straight-line segment, and all points of those straight-line segments lie entirely within the object. A star is a representative 2D example of this class of objects. All convex objects are also star-shaped objects (but not vice versa).

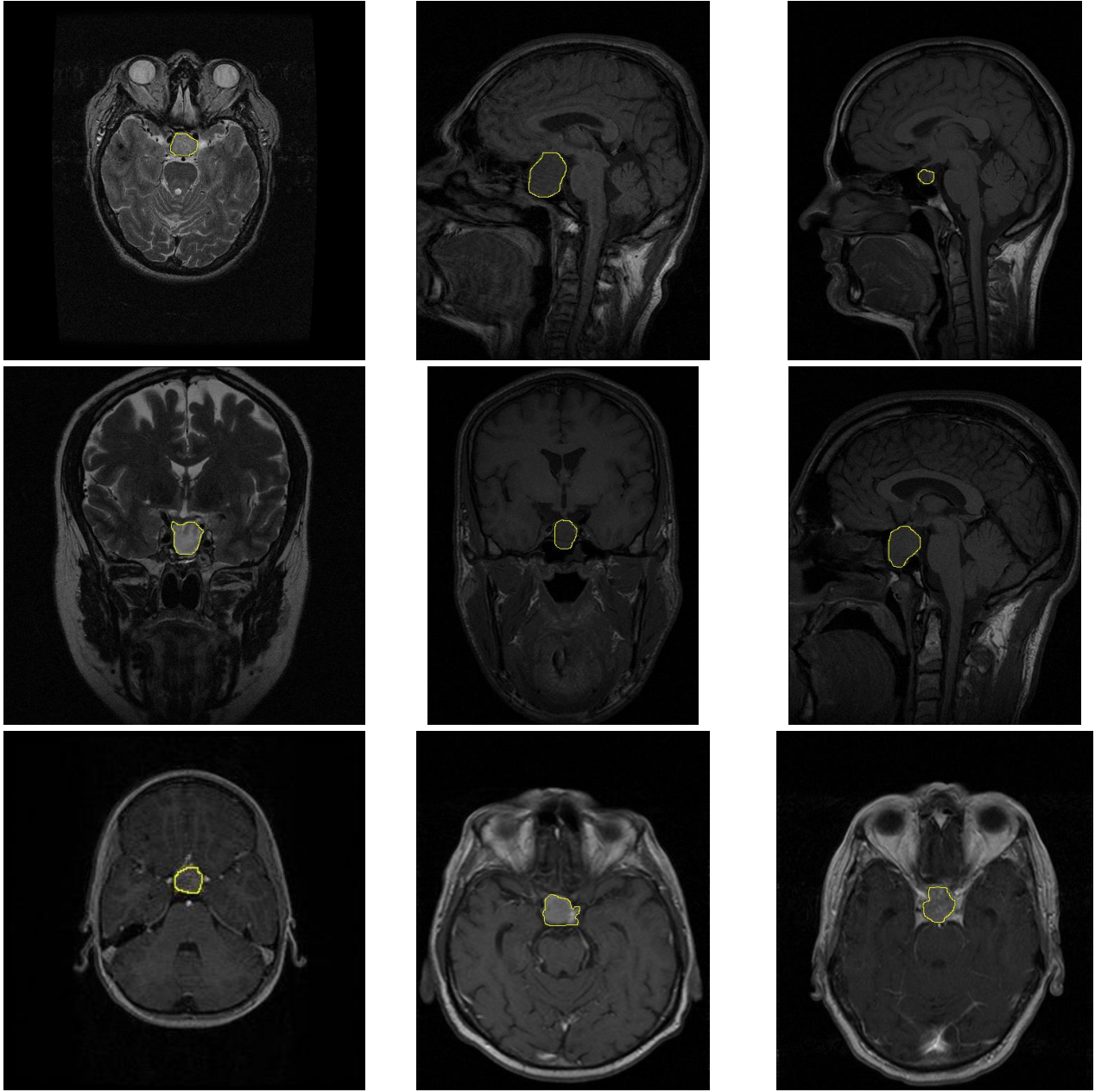

Fig. 2: User initialization (yellow outline) superimposed onto nine magnetic resonance imaging acquisitions of pituitary adenoma datasets that have been used for evaluation of the proposed method (the tenth is shown in Figure 1).

The following steps are performed iteratively:

1. Split polyhedron edges that are S times longer than mean voxel spacing (geometrical mean of spacing in X, Y and Z direction). We used S=2.95. This parameter influences the smoothness of the mesh, as well as precision and speed of the segmentation, and it is thus a tuning parameter. S should not be smaller than the voxel spacing, since this would waste computing power and counteract the smoothness enforcement (see steps 3 and 4).

2. Compute per-vertex surface normals and curvature estimates. Since we use a relatively large S, we can estimate the curvature relying on the 1-neighborhood of the vertex. With smaller S, the curvature estimate should rely on larger vertex neighborhoods.

3. Move vertices outwards (inflate the mesh):

   a. Surface orientation: Take into consideration the cosine of the angle (φ) between center-vertex vector ($\vec{d_{cv}}$) and surface normal vector ($\vec{n_v}$) for each vertex. The greater the angle, the lower the inflation speed. This slows down the inflation when the mesh starts adapting to the shape of the boundary.

   b. Surface curvature: The higher the curvature, the lower the inflation speed, thus the inflation speed is lower for vertices near feature points (ridges, valleys, peaks and dents).

   c. If a vertex can be moved, it is moved in the direction of center-vertex vector (thus maintaining the star shape). The vertex cannot be moved if the destination voxel has intensity outside the range of interest (contents of the user-drawn boundary). Otherwise, if the destination voxel intensity is higher than or close to the maximum intensity this vertex has encountered recently, the vertex can be displaced. This favors more common boundaries with lower intensity surrounding tissues. The displacement amount is adjusted by inflation speed factor (Figures 3 and Figure 4).

4. Smooth the surface of the polyhedron slightly. This is required to overcome noisy voxels, which would otherwise prevent inflation of the mesh beyond them, even if they are in the middle of the tumor. This way we do not need to denoise the input dataset, which is a computationally expensive operation. We also know that the surface of the tumor is smooth.

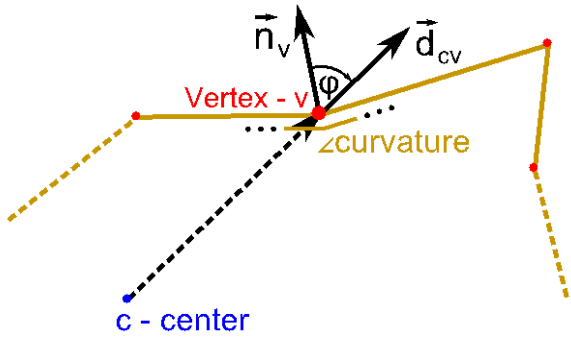
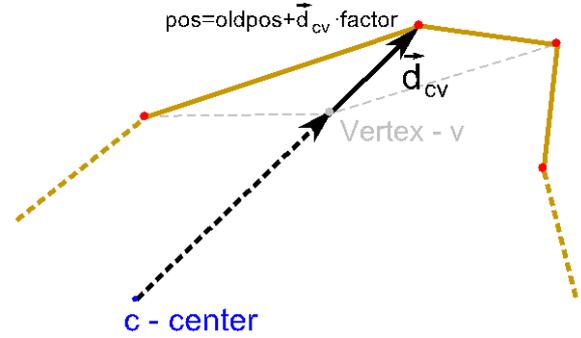

Fig. 3: Mesh properties relevant for movement of each vertex.   Fig. 4: Effect of moving a vertex outwards.

## 4. RESULTS AND CONCLUSION

The presented approach was realized in C++ and the automatic segmentation in our implementation took about one second per dataset (about 30 seconds including the time it takes to locate the file, draw an outline, execute segmentation and extract a marching cubes isosurface) on an Intel Core i7-920 CPU (2.66 GHz) with a GeForce 8800GTX graphics card on Windows7 x64. The evaluation set consisted of eight T1 weighted and two T2 weighted images. 9 images had a slice resolution of 512x512 (and between 13 and 80 slices), and one had a resolution of 256x256x160. Note that the increase of the size in image regions unrelated to the tumor (scan of entire head instead of only the segment that contains the tumor) does not affect the segmentation speed.

To evaluate the approach, neurological surgeons with several years of experience in the resection of brain tumors performed manual slice-by-slice segmentation of ten pituitary adenomas. Afterwards, the manual segmentations were compared with the segmentation results of the proposed method via the Dice Similarity Coefficient (DSC)[13, 14]. The Dice Similarity Coefficient is the relative volume overlap between A and R, where A and R are the binary masks from the automatic (A) and the reference (R) segmentation. V(•) is the volume (in cm$^3$) of voxels inside the binary mask, by means of counting the number of voxels, then multiplying with the voxel size:

$$DSC = \frac{2 \cdot V(A \cap R)}{V(A) + V(R)} \quad (1)$$

The average DSC for all datasets was 75.92%±7.24% (minimum 63.74 and maximum 86.08, see Table 1 for details). Compared to a manual segmentation that took, on the average, 3.91±0.54 minutes, the segmentation in our implementation required about one second.

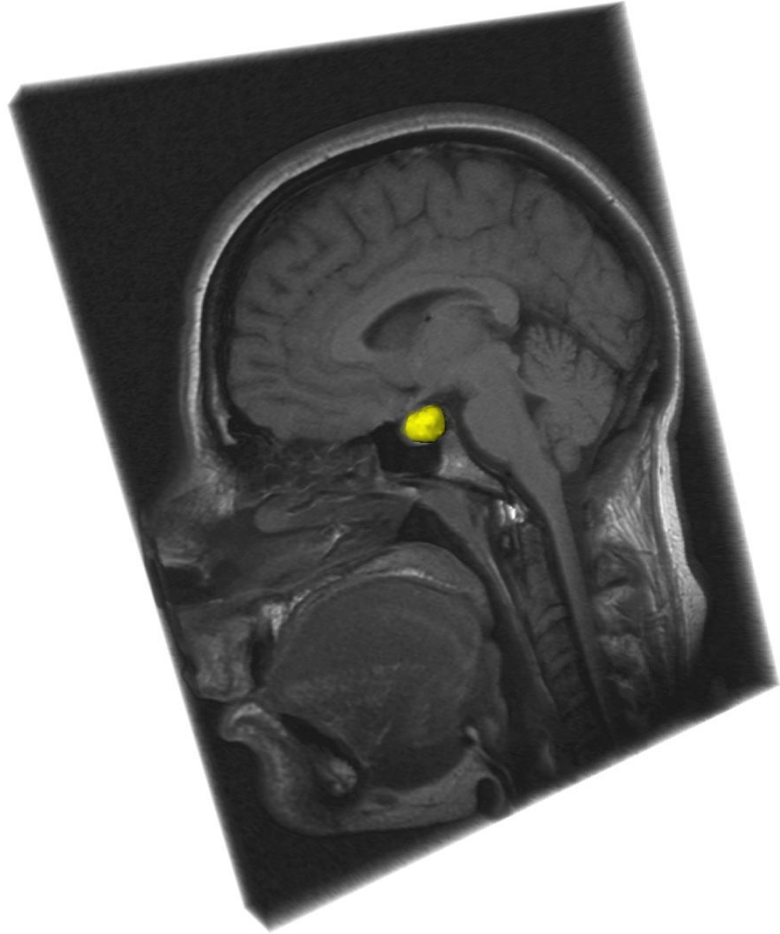

Fig. 5: Resulting segmentation superimposed into semi-transparent cutout of the dataset, which corresponds to Figure 1.

|  | Volume of tumor (cm$^3$) | | Number of voxels | | DSC (%) | manual seg. time (min) |
| --- | --- | --- | --- | --- | --- | --- |
|  | manual | algorithm | manual | algorithm |  |  |
| **min** | 0.84 | 0.6 | 4492 | 3525 | 63.74 | 3 |
| **max** | 15.57 | 13.05 | 106151 | 88986 | 86.08 | 5 |
| $\mu(\pm\sigma)$ | 6.30 ± 4.07 | 4.69 ± 3.58 | 47462.7 | 34317.8 | 75.92 ± 7.24 | 3.91 ± 0.54 |

Table 1: Summary of results: min., max., mean $\mu$ and standard deviation $\sigma$ for ten pituitary adenomas.

Figures 1 and 2 show the user initializations (yellow outline) that have been superimposed onto a slice of the magnetic resonance imaging acquisitions of all ten pituitary adenoma datasets. The shown slices are the ones user selected as approximately central to the tumor. The segmentation results for the ten pituitary adenoma datasets are presented in Figures 5 and 6, by visualizing them as a three-dimensional closed surface model (yellow) faded into the corresponding MRI dataset (the order in Figure 6 corresponds to Figure 2).

In conclusion, the proposed method can be used to augment the manual segmentation, reducing the work to contour corrections where necessary. It imposes no special preprocessing requirements, executes very quickly on modern hardware, and provides decent results. However, there are several areas of future work. For example, some parameter specifications of the proposed algorithm can be automated and we want to exchange the contour initialization with a single user-defined seed point that is placed in the pituitary adenoma. Additionally, we plan a comparison of the segmentation results with a recently introduced approach for spherically and elliptically shaped objects[15, 16] that has also been used for pituitary adenoma segmentation[10].

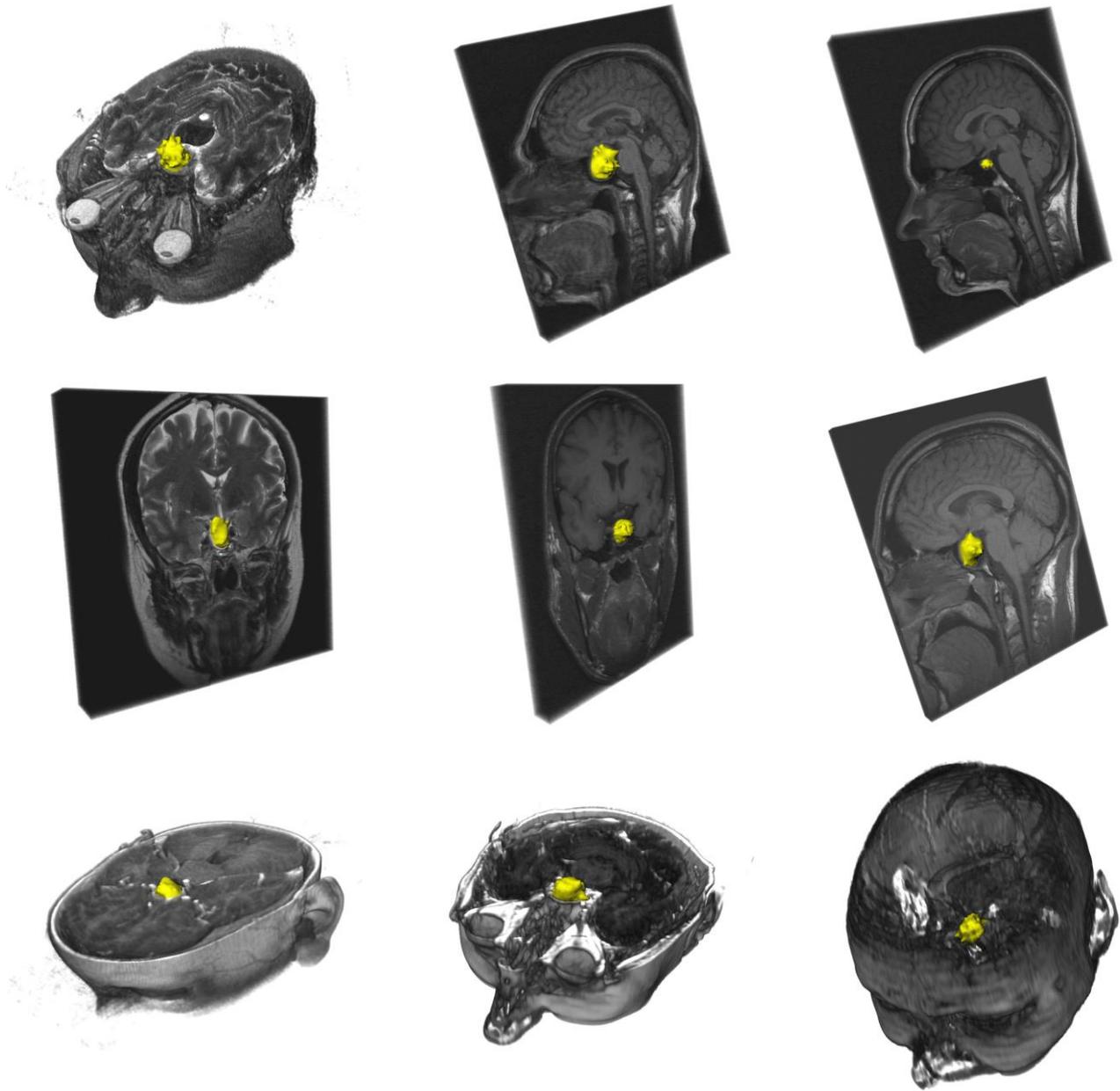

Fig. 6: Resulting segmentations as 3D models (yellow) superimposed into nine magnetic resonance imaging acquisitions of pituitary adenomas (the tenth is shown in Figure 5). The order corresponds to the user initializations presented in Figure 2.

## ACKNOWLEDGMENTS

The authors would like to thank German Academic Exchange Service (DAAD) for funding (International Postgraduate Program at Zentrum für Sensor Systeme in Siegen). Furthermore the authors would like to thank Fraunhofer MeVis in Bremen, Germany, for their collaboration and especially Horst K. Hahn for his support.